\documentclass{article}

\PassOptionsToPackage{numbers, compress}{natbib}

 \usepackage[preprint]{nips_2018}

\usepackage[utf8]{inputenc} 
\usepackage[T1]{fontenc}    
\usepackage{hyperref}       
\usepackage{url}            
\usepackage{booktabs}       
\usepackage{amsfonts}       
\usepackage{nicefrac}       
\usepackage{microtype}      

\usepackage{graphicx}
\graphicspath{ {Diagrams/} }
\usepackage{subcaption}
\usepackage{wrapfig}  
\usepackage{listings}
\usepackage{amsmath}
\usepackage{amssymb}
\usepackage[retainorgcmds]{IEEEtrantools}
\usepackage[inline]{enumitem}

\newcommand{\website}{https://untrix.github.io/i2l/}
\newcommand{\implementation}{https://github.com/untrix/im2latex}
\newcommand{\suppmat}{appendix}
\newcommand{\Suppmat}{Appendix}

\title{Teaching Machines to Code:\\ Neural Markup Generation with Interpretable Attention}

\author{
	Sumeet S. Singh \\
	Independent Researcher \\
	Saratoga, CA 95070 \\
	sumeet@singhonline.info
}

\begin{document}
\hyphenation{image video speech text synthesis description handwriting recognition}
\newcommand{\xtm}{\ensuremath{\boldsymbol{x}_t}}
\newcommand{\htm}[1][t]{\ensuremath{\boldsymbol{h}_{#1} }}
\newcommand{\ctm}[1][t]{\ensuremath{\boldsymbol{c}_{#1} }}
\newcommand{\ztm}[1][t]{\ensuremath{\boldsymbol{z}_{#1} }}
\newcommand{\itm}[1][t]{\ensuremath{\boldsymbol{i}_{#1} }}
\newcommand{\otm}[1][t]{\ensuremath{\boldsymbol{o}_{#1} }}
\newcommand{\ftm}[1][t]{\ensuremath{\boldsymbol{f}_{#1} }}
\newcommand{\wtmat}[2]{W_{#1 #2}}

\maketitle

\begin{abstract}
	We present a neural transducer model with visual attention that learns to generate \LaTeX  ~markup of a real-world math formula given its image. Applying sequence modeling and transduction techniques that have been very successful across modalities such as natural language, image, handwriting, speech and audio; we construct an image-to-markup model that learns to produce syntactically and semantically correct \LaTeX ~markup code over 150 words long and achieves a BLEU score of 89\%; improving upon the previous state-of-art for the Im2Latex problem. We also demonstrate with heat-map visualization how attention helps in interpreting the model and can pinpoint (localize) symbols on the image accurately despite having been trained without any bounding box data.
\end{abstract}

\section{Introduction}
\label{intro}
In the past decade, deep neural network models based on RNNs\footnote{Recurrent Neural Network.}, CNNs\footnote{Convolutional Neural Networks and variants such as dilated CNNs \cite{Yu2015MultiScaleCA}.} and `attention' \cite{Vaswani2017AttentionIA} have been shown to be very powerful sequence modelers and transducers. Their ability to model joint distributions of real-world data has been demonstrated through remarkable achievements in a broad spectrum of generative tasks such as;
\begin{itemize*}
	\item[{image synthesis}] \cite{Oord2016PixelRN, Oord2016ConditionalIG, Salimans2017PixelCNNIT, Theis2015GenerativeIM}, 
	\item[{image description}] \cite{Karpathy2015DeepVA, Xu2015ShowAA, Johnson2016DenseCapFC, DBLP:journals/corr/PedersoliLSV16, Vinyals2015ShowAT}, 
	\item[{video description}] \cite{Donahue2015LongtermRC}, 
	\item[{speech and audio synthesis}] \cite{Oord2016WaveNetAG}, 
	\item[{handwriting recognition}] \cite{Graves2008OfflineHR, Bluche2016JointLS}, 
	\item[{handwriting synthesis}] \cite{Graves2013GeneratingSW}, 
	\item[{machine translation}] \cite{Cho2014LearningPR, Bahdanau2014NeuralMT, Kalchbrenner2016NeuralMT, Sutskever2014SequenceTS}, 
	\item[{speech recognition}] \cite{Graves2006ConnectionistTC, DBLP:journals/corr/ChanJLV15, DBLP:journals/corr/abs-1303-5778},
	\item[{etc.}] \cite{Graves2008SupervisedSL, Vaswani2017AttentionIA}
\end{itemize*}

One class of sequence models employ the so-called \textbf{\emph{encoder-decoder}} \cite{Cho2014LearningPR} or \textbf{\emph{sequence-to-sequence}}  \cite{Sutskever2014SequenceTS} architecture, wherein an \emph{encoder} encodes a source sequence into feature vectors, which a \emph{decoder} employs to produce the target sequence. The source and target sequences may either belong to the same modality (e.g. in machine translation use-cases) or different modalities (e.g. in image-to-text, text-to-image, speech-to-text); the encoder / decoder sub-models being constructed accordingly. The entire model is trained end-to-end using supervised-learning techniques. In recent years, this architecture has been augmented with an \textbf{\emph{attention and alignment }} model which selects a subset of the feature vectors for decoding. It has been shown to help with longer sequences \cite{Bahdanau2014NeuralMT, DBLP:journals/corr/LuongPM15}. Among other things, this architecture has been used for image-captioning \cite{Xu2015ShowAA}.
In our work we employ a encoder-decoder architecture with attention, to map images of math formulas into corresponding \LaTeX ~markup code.
The contributions of this paper are:
\begin{enumerate*}[label=\arabic*)]
	\item Solves the Im2Latex problem\footnotemark[100] and improves over the previous best reported BLEU score by 1.27\% BLEU, 
	\item Pushes the boundaries of the neural encoder-decoder architecture with visual attention, 
	\item Analyses variations of the model and cost function. Specifically we note the changes to the base model \cite{Xu2015ShowAA} and what impact those had on performance,  
	\item Demonstrates the use of attention visualization for model interpretation and
	\item Demonstrates how attention can be used to localize objects (symbols) in an image despite having been trained without bounding box data.
\end{enumerate*}

\subsection{The \textsc{Im2Latex} problem}
\label{the_problem}
The \href{https://openai.com/requests-for-research/#im2latex}{\textsc{Im2Latex} Problem} is a request for research
proposed by \href{https://openai.com}{OpenAI}. The challenge is to build a Neural Markup
Generation model that can be trained end-to-end to generate the \LaTeX ~markup of a math formula given its image.
Data for this problem 
was produced by rendering single-line real-world \LaTeX  ~formulas obtained from the \href{http://www.cs.cornell.edu/projects/kddcup/datasets.html}{KDD Cup 2003 dataset}. The resulting grayscale images were used as the input samples while the original markup was used as the label/target sequence.
\lstset{basicstyle=\tiny}
\begin{figure}[!h]
	\begin{displaymath}
	S _ { 0 } = \sum _ { l } \frac { 1 } { 2 \Delta _ { l } ^ { 2 } } \mathrm { T r } \, \phi _ { l } ^ { a } \phi _ { - l } ^ { a } + \sum _ { l } \frac { 1 } { 2 \epsilon _ { l } ^ { 2 } } \mathrm { T r } \, f _ { l } ^ { a } f _ { - l } ^ { a } + \sum _ { r } \frac { 1 } { g _ { r } } \mathrm { T r } \, \bar { \psi } _ { r } ^ { a } \psi _ { r } ^ { a } \, .
	\end{displaymath}
	\lstinline|S _ { 0 } = \sum _ { l } \frac { 1 } { 2 \Delta _ { l } ^ { 2 } } \mathrm { T r } \, \phi _ { l } ^ { a } \phi _ { - l } ^ { a } + \sum _ { l } \frac { 1 } { 2 \epsilon _ { l } ^ { 2 } } \mathrm { T r } \, f _ { l } ^ { a } f _ { - l } ^ { a } + \sum _ { r } \frac { 1 } { g _ { r } } \mathrm { T r } \, \bar { \psi } _ { r } ^ { a } \psi _ { r } ^ { a } \, .| \\
		
	\lstinline|S _ { 0 } = \sum _ { l } { \frac { 1 } { 2 \Delta _ { l } ^ { 2 } } } \mathrm { T r } \, \phi _ { l } ^ { a } \phi _ { - l } ^ { a } + \sum _ { l } { \frac { 1 } { 2 \epsilon _ { l } ^ { 2 } } } \mathrm { T r } \, f _ { l } ^ { a } f _ { - i } ^ { a } + \sum _ { r } { \frac { 1 } { g _ { r } } } \mathrm { T r } \, \psi _ { r } ^ { a } \psi _ { r } ^ { a } \, .|
	\caption{A training sample: At the top is the input image $\boldsymbol{x}$, middle the target sequence $\boldsymbol{y}$ ($\tau = 145$) and bottom the predicted sequence $\boldsymbol{\hat{y}}$ (${\tau} = 148$). Each space-separated word in $\boldsymbol{y}$ and $\boldsymbol{\hat{y}}$ $\in$ $V$ }
	\label{fig-sample}
\end{figure}
Each training/test sample (Figure \ref{fig-sample}) is comprised of an input image  $\boldsymbol{x}$ and a corresponding target \LaTeX -sequence $\boldsymbol{y}$ of length $\tau$. Each word $\boldsymbol{y}$ of the target sequence, belongs to the vocabulary of the dataset plus two special tokens: beginning-of-sequence <bos> and end-of-sequence <eos>. 
Denoting image dimensions as $H_I, W_I$ and $C_I$ and the vocabulary as a set $V$ of $K$ words, we represent \smash{$\boldsymbol{x} \in \mathbb{R}^{H_I \times W_I \times C_I}$}, \smash{$V := \lbrace \text{\LaTeX ~tokens}, \text{<eos>}, \text{<bos>} \rbrace  ; |V| = K $} and \smash{$\boldsymbol{y} := (  \boldsymbol{y}_1, \ldots, \boldsymbol{y}_{\tau} ) ; \quad \boldsymbol{y}_t \in \lbrace 1, \ldots , K \rbrace$}.
The task is to generate markup that a \LaTeX ~compiler will render back to the original image. Therefore, our model needs to generate syntactically and semantically correct markup, by simply `looking' at the image: i.e. it should jointly model vision and language.

\section{Image to markup model}
\begin{figure}[!h]
	\begin{subfigure}{0.5\textwidth}
		\includegraphics{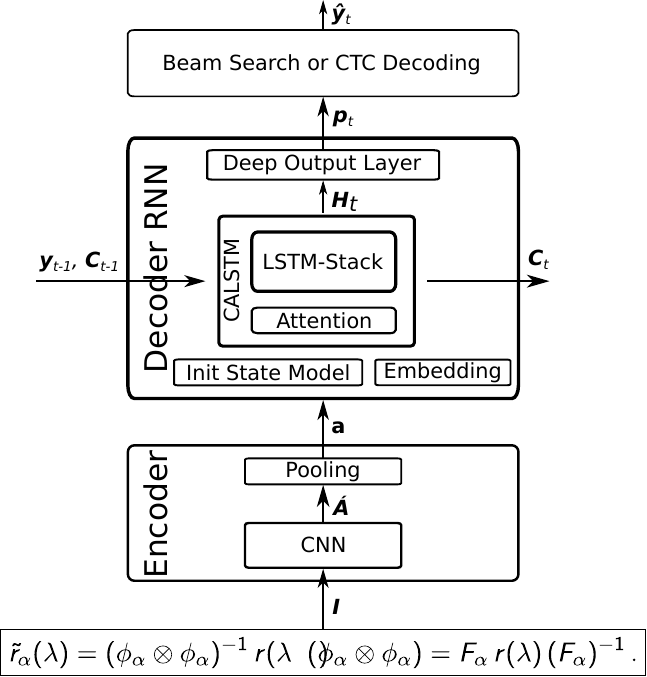}
		\centering
		\caption{}
		\label{fig-i2l-brief}
	\end{subfigure}
	\begin{subfigure}{0.5\textwidth}
		\includegraphics{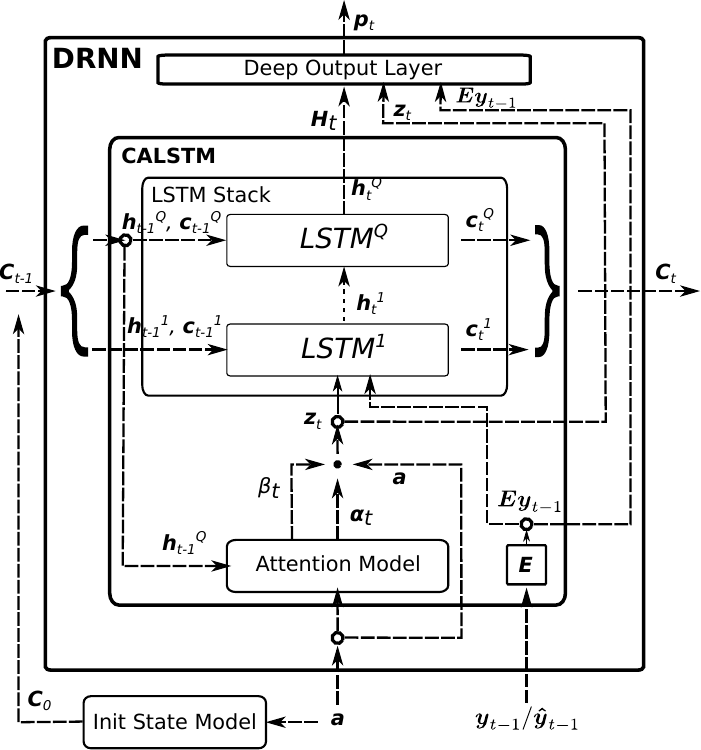}
		\centering
		\caption{}
		\label{fig-decoder}
	\end{subfigure}
	\caption[Model]{(a) Model outline showing major parts of the model. Beam search decoder is only used during inferencing, not training. LSTM-Stack and Attention model jointly form a Conditioned Attentive LSTM stack (CALSTM) which can itself be stacked. (b) Expanded view of Decoder RNN showing its sub-models. There are three nested RNN cells in all: The decoder RNN (DRNN) at the top level, nesting the CALSTM which nests the LSTM-Stack. The Init Model does not participate in recurrence, therefore its is shown outside the box.}
\end{figure}

Our model (Figure \ref{fig-i2l-brief}) has the same basic architecture as \cite{Xu2015ShowAA} (which we call our baseline model) in the way the encoder, decoder and a visual attention interact. However there are significant differences in the sub-models which we notate in the remainder of this paper and in the \suppmat.

\subsection{Encoder}
\label{encoder-brief}
\begingroup
\setlength{\columnsep}{0pt}%
All images are standardized to a fixed size by centering and padding with white pixels. Then they are linearly transformed (whitened) to lie in the range [-0.5,0.5]. A deep CNN then encodes the whitened image into a \emph{visual feature grid} ${\boldsymbol{\acute{A}}}$, having \smash{$\acute{H} \times \acute{W}$} (i.e. height $\times$ width) \emph{visual feature vectors} {$\boldsymbol{a}_{(\acute{h}, \acute{w})} \in \mathbb{R}^{\acute{D}}$}.
\begin{wrapfigure}[6]{r}[0pt]{0.45\textwidth}
	\vspace{-5pt}
	\begin{IEEEeqnarray}{c}
		\boldsymbol{A} :=
		\begin{bmatrix}
			\boldsymbol{a}_{(1,1)} & \ldots & \boldsymbol{a}_{(1,W)} \\
			\vdots & \vdots & \vdots \\
			\boldsymbol{a}_{(H,1)} & \ldots & \boldsymbol{a}_{(H,W)} \\
		\end{bmatrix} 
	\end{IEEEeqnarray}
\end{wrapfigure}
The visual feature vectors are then concatenated (pooled) together in strides of shape $\smash{[S_H, S_W]}$; 
begetting \emph{pooled feature vectors} $\boldsymbol{a}_{(h,w)} \in \mathbb{R}^D$, where ${D = \acute{D} \cdot S_H \cdot S_W}$. The resulting feature map $\boldsymbol{A}$, has a correspondingly shrunken shape $[H, W]$; where $H = \acute{H} / S_H$ and $W = S_W / \acute{W}$.

Each pooled feature vector can be viewed as a rectangular window into the image, bounded by its receptive field.\footnote{Neighboring regions overlap but each region is distinct overall.} The idea behind this is to partition the image into spatially localized regional encodings and setup a decoder architecture (Section \ref{decoder-brief}) that selects/emphasizes only the relevant regions at each time-step $t$, while filtering-out/de-emphasizing the rest. \citealt{Bahdanau2014NeuralMT} showed that such piecewise encoding enables modeling longer sequences as opposed to models that encode the entire input into a single feature vector \cite{Sutskever2014SequenceTS, Cho2014LearningPR}.
\footnote{That said, \citealt{Bahdanau2014NeuralMT} employ a bidirectional-LSTM \cite{Graves2008SupervisedSL} encoder whose receptive field does encompass the entire input anyway! ({Although that does not necessarily mean that the bi-LSTM will encode the entire image}). Likewise \citealt{Deng2017ImagetoMarkupGW} who also solve the \textsc{Im2Latex} problem also employ a bi-directional LSTM stacked on top of a CNN-encoder in order to get full view of the image. In contrast, our visual feature vectors hold only spatially local information which we found are sufficient to achieve good accuracy. This is probably owing to the nature of the problem; i.e. transcribing a one-line math formula into \LaTeX sequence requires only local information at each step.}
Pooling allows us to construct encoders with different receptive field sizes. We share results of two such models:
\begin{itemize*}
	\item[\textsc{i2l-nopool}] with no feature pooling and pooled feature grid shape [4,34] and 
	\item[\textsc{i2l-strips}] having stride [4,1] and pooled feature grid shape [1,34].
\end{itemize*}	
Finally, for convenience we represent $\boldsymbol{A}$ as a flattened sequence $\boldsymbol{a}$ (Equation \ref{eqn-a}). See the \suppmat ~for more details.
\begin{equation}
\boldsymbol{a} := \left( \boldsymbol{a}_1, \ldots , \boldsymbol{a}_L  \right); \; \boldsymbol{a}_{l} \in \mathbb{R}^{D}; \;  l = H(h-1) + w; \; L = H W \label{eqn-a}
\end{equation}

\endgroup

\subsection{Decoder}
\label{decoder-brief}
\begin{wrapfigure}[7]{r}{0.4\textwidth}
	\vspace{-22pt}
	\begin{IEEEeqnarray}{rC;l}
		\boldsymbol{p}_t & : & \lbrace 1, \ldots , K \rbrace \rightarrow [0,1]  \nonumber \\
		\boldsymbol{y_t} & \sim & \boldsymbol{p}_t  \nonumber \\
		\boldsymbol{p}_t(\boldsymbol{y_t}) & := &  P_r(\boldsymbol{y}_{t} | \boldsymbol{y}_{<t}, \boldsymbol{a}) \label{eqn-decoder-defns}
	\end{IEEEeqnarray}
		\begin{IEEEeqnarray}{rCl}
			P_r (\boldsymbol{y}|\boldsymbol{a})  = & \prod_{t=1}^{\tau} \boldsymbol{p}_t \left(  \boldsymbol{y}_t \right) \label{eqn-sequence-prob}
		\end{IEEEeqnarray}
\end{wrapfigure}
The decoder is a language modeler and generator. It is a Recurrent Neural Network (DRNN in Figure \ref{fig-decoder}) that models the discrete probably distribution $\boldsymbol{p}_t$, of the output word $\boldsymbol{y}_{t}$, conditioned on the sequence of previous words $\boldsymbol{y}_{<t}$ and relevant regions of the encoded image $\boldsymbol{a}$\footnote{This is now a very standard way to model sequence (sentence) probabilities in neural sequence-generators. See \cite{Sutskever2014SequenceTS} for example.} (Equations \ref{eqn-decoder-defns}). Probability of the entire output sequence $\boldsymbol{y}$ given image $\boldsymbol{a}$ is therefore given by Equation \ref{eqn-sequence-prob}.

The DRNN receives the previous word $\boldsymbol{y}_{t-1}$ and encoded image $\boldsymbol{a}$ as inputs. In addition, it maintains an internal state $\boldsymbol{C}_t$ that propagates information (features) extracted from an initial state, the output sequence unrolled thus far and image regions attended to thus far (Equation \ref{eqn-rnn}).
	\begin{figure}
	\begin{IEEEeqnarray}{rCcCl}
		\text{DRNN} & : & \lbrace \boldsymbol{a}; \,  \boldsymbol{y}_{t-1}; \, \boldsymbol{C}_{t-1} \rbrace & \rightarrow & \lbrace  \boldsymbol{p}_t; \, \boldsymbol{C}_{t} \rbrace \label{eqn-rnn}
	\end{IEEEeqnarray}
\end{figure}
It is as complex model, comprised of the following sub-models (Figure \ref{fig-decoder}):
\begin{enumerate*}[label=\arabic*)]
	\item A {LSTM-Stack} \cite{Hochreiter:1997:LSM:1246443.1246450} responsible for memorizing $\boldsymbol{C}_t$ and producing a recurrent activation $\boldsymbol{H}_{t}$,
	\item A Visual {attention} and alignment model responsible for selecting relevant regions of the encoded image for input to the LSTM-Stack, \footnote{The LSTM-Stack and Visual Attention and Alignment model jointly form a Conditioned Attentive LSTM (CALSTM); $\boldsymbol{H}_t$ and $\boldsymbol{C}_t$ being its activation and internal state respectively. Our source-code implements the CALSTM as a RNN cell which may be used as a drop-in replacement for a RNN cell.}
	\item A {Deep Output} Layer \cite{Pascanu2013HowTC} that produces the output probabilities $\boldsymbol{p}_t$,
	\item {Init Model}: A model that generates the initial state $\boldsymbol{C}_{0}$ and
	\item An {embedding} matrix $\boldsymbol{E}$ (learned by training) that transforms $\boldsymbol{y}_t$ into a dense representation $\in \mathbb{R}^m$.
\end{enumerate*}

\subsubsection{Inferencing}
After the model is trained, the output sequence is generated by starting with the word `bos' and then repeatedly sampling from $\boldsymbol{p}_t$ until <eos> is produced. The sequence of words thus sampled is the predicted sequence: $\boldsymbol{\hat{y}} := ( \boldsymbol{\hat{y}}_1, \ldots, \boldsymbol{\hat{y}}_{\hat{\tau}} ) \, ; \, \boldsymbol{\hat{y}}_{t} \in \mathbb{R}^K$.  For this procedure we use beam search decoding \cite{Graves2008SupervisedSL} with a beam width of 10. Figure \ref{fig-sample} shows an example predicted sequence and Figures \ref{fig-good-preds} and \ref{fig-bad-preds} show examples of predictions rendered into images by a \LaTeXe ~compiler.


\subsubsection{Visual attention and alignment model}
\label{attention}
\begin{wrapfigure}[5]{r}{0.45\textwidth}
	\vspace{-23pt}
	\begin{IEEEeqnarray}{rCl}
		\boldsymbol{\alpha}_t &:= & \left( \alpha_{t,1} , \ldots , \alpha_{t,L}  \right) \; \Big|
		\begin{array}{l}
			\scriptstyle{0 \leq \alpha_{t,l} \leq 1} \\
			\scriptstyle{\sum_{l}^{L} \alpha_{t,l} = 1}
		\end{array} \label{eqn-alpha} \\
		\boldsymbol{\alpha}_t &= & f_{att} \left( \boldsymbol{a} ; \, \boldsymbol{H}_{t-1} \right) \label{eqn-fatt} \\
		\boldsymbol{z}_t & = & \boldsymbol{\alpha}_t \boldsymbol{a}^\top \label{eqn-z}
	\end{IEEEeqnarray}
\end{wrapfigure}
As previously alluded, the decoder soft selects/filters relevant (encoded) image regions at each step. This is implemented via. a `soft attention' mechanism\footnote{`Soft' attention as defined by \citet{Xu2015ShowAA} and originally proposed by \citet{Bahdanau2014NeuralMT}.} which computes a weighted sum $\boldsymbol{z}_t$ of the pooled feature vectors $\boldsymbol{a}_l$. The visual attention model $f_{att}$, computes the weight distribution $\boldsymbol{\alpha}_t$ (Equations \ref{eqn-alpha}, \ref{eqn-fatt} and \ref{eqn-z}).
$f_{att}$ is modeled by an MLP (details in the \suppmat).
\begin{figure}
	\includegraphics[width=0.5\textwidth]{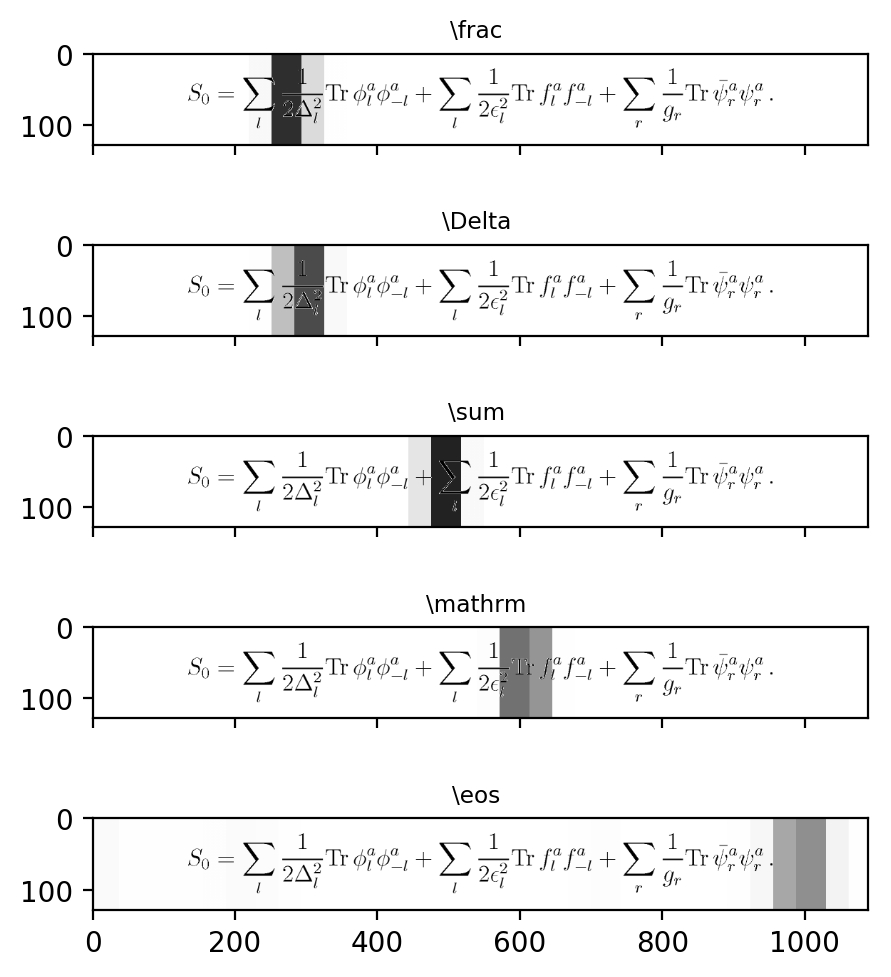}
	\includegraphics[width=0.5\textwidth]{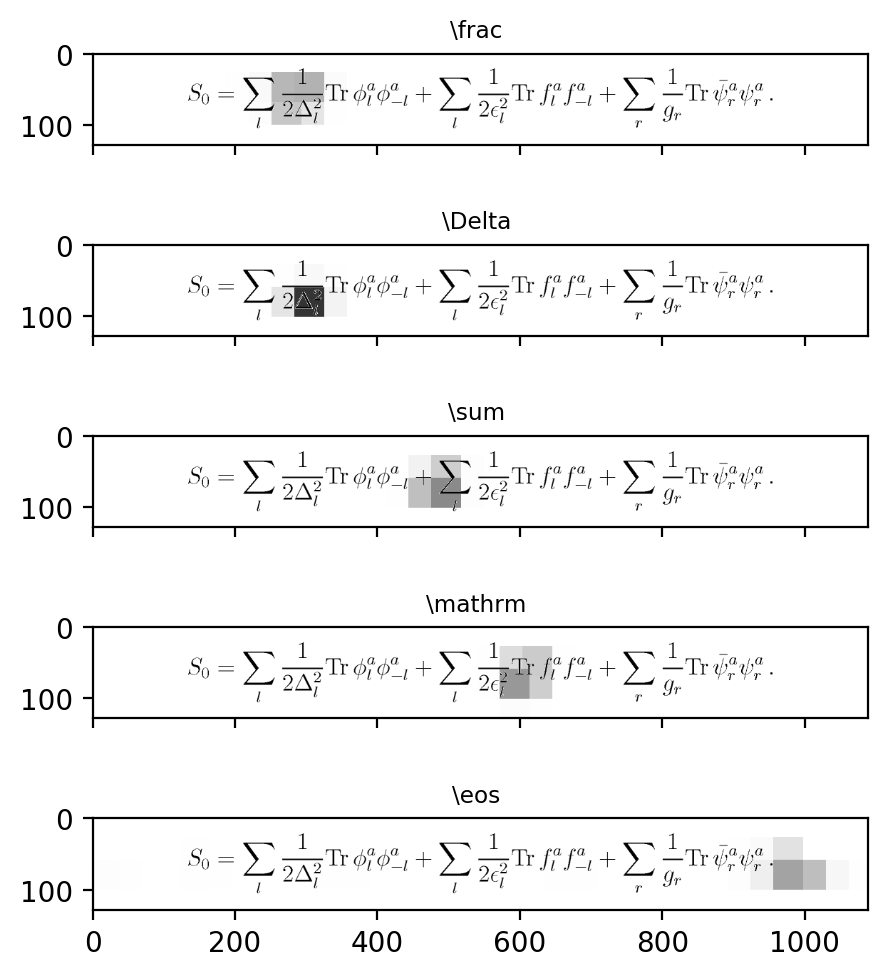}
	\caption[Visual Attention]{Focal-regions learnt by the attention model: to the left by I2L-STRIPS and to the right by I2L-NOPOOL. Image darkness is proportional to $\boldsymbol{\alpha}_t$. Notice how $\boldsymbol{\alpha}_t$ concentrates on the image region corresponding to the output word (shown above the image). The \textbackslash frac command starts a fraction, \textbackslash mathrm sets a font and \textbackslash eos is the <eos> token.}
	\label{fig-att}
\end{figure}
While it is a possible for $\boldsymbol{\alpha}_t$ to end up uniformly distributed over $( \boldsymbol{a}_1  \ldots  \boldsymbol{a}_L )$, in practice we see a unimodal shape with most of the weight concentrated on 1-4 neighborhood (see Figure \ref{fig-att}) around the mode. We call this neighborhood the \emph{focal-region} - i.e. the focus of attention. In other words we empirically observe that the attention model's focus is `sharp'; converging towards the `hard attention' formulation described by \citet{Xu2015ShowAA}. Also note that (Figure \ref{fig-att}), the attention model is able to utilize the extra granularity available to it in the I2L-NOPOOL case and consequently generates much sharper focal-regions than I2L-STRIPS.

Furthermore, the model aligns the focal-region with the output word and thus scans text on the image left-to-right (I2L-STRIPS) or left-right and up-down (I2L-NOPOOL) just like a person would read it (Figure \ref{fig-att}). We also observe that it doesn't focus on empty margins of the image except at the first and last (<eos>) steps which is quite intuitive for determining the beginning or end of text.

\subsubsection{LSTM stack}
\label{lstm-stack}
\begin{wrapfigure}[7]{r}[0pt]{0.45\textwidth}
	\vspace{-15pt}
	\begin{IEEEeqnarray*}{rCl}
		LSTM^q &:& \lbrace \boldsymbol{x}^{q}_{t}; \boldsymbol{h}^{q}_{t-1} ; \boldsymbol{c}^{q}_{t-1} \rbrace \rightarrow  \lbrace  \boldsymbol{h}^q_{t} ; \boldsymbol{c}^q_{t} \rbrace  \\
		&&1 \leq q \leq Q  \; ; \; \boldsymbol{h}_t^q, \boldsymbol{c}_t^q \in \mathbb{R}^n  \\
		\xtm^q &=& \boldsymbol{h}^{q-1}_{t} \quad ; q \ne 1 \IEEEyesnumber \label{eqn-lstm-stack} \\
		\xtm^1 &=& \lbrace \boldsymbol{z}_t; \, \boldsymbol{Ey}_{t-1}  \rbrace 
	\end{IEEEeqnarray*}
\end{wrapfigure}
The core sequence generator of the DRNN is a multilayer LSTM \cite{Graves2013GeneratingSW} (Figure \ref{fig-decoder}). Our LSTM cell implementation follows \citet{DBLP:journals/corr/abs-1303-5778}. The LSTM cells are stacked in a multi-layer configuration \cite{Zaremba2014RecurrentNN, Pascanu2013HowTC} as in Equation \ref{eqn-lstm-stack}.
$LSTM^q$ is the LSTM cell at position $q$ with $\xtm^q$, $\boldsymbol{h}^q_t$ and $\boldsymbol{c}^q_t$ being its input, hidden activation and cell state respectively. $LSTM^1$ receives the stack's input: soft attention context $\ztm$ and previous output word $\boldsymbol{Ey}_{t-1}$. $LSTM^Q$ produces the stack's output $\boldsymbol{H}_t = \htm^Q$, which is sent up to the Deep Output Layer. Accordingly, the stack's activation ($\boldsymbol{H}_t$) and state ($\boldsymbol{C}_t$) are defined as: $\boldsymbol{H}_t = \boldsymbol{h}^Q_{t}$ and $\boldsymbol{C}_t := ( \boldsymbol{c}^1_{t}, \ldots , \boldsymbol{c}^Q_{t}, \, \boldsymbol{h}^1_{t}, \ldots , \boldsymbol{h}^Q_{t} ) $.
We do not use skip or residual connections between the cells. Both of our models have two LSTM layers with $n = 1500$. Further discussion and details of this model can be found in the \suppmat.

\subsubsection{Deep output layer}
We use a Deep Output Layer \cite{Pascanu2013HowTC} to produce the final output probabilities: $\boldsymbol{p}_t = f_{out}(\boldsymbol{H}_t; \, \boldsymbol{z}_t; \, \boldsymbol{Ey}_{t-1})$.
$f_{out}$ is modeled by an MLP. Note that the output layer receives skip connections from the LSTM-Stack input (Equation \ref{eqn-lstm-stack}). Details of this model can be found in the \suppmat.

\subsubsection{Init model}
\begin{wrapfigure}[11]{r}{0.25\textwidth}
	\centering
	\includegraphics{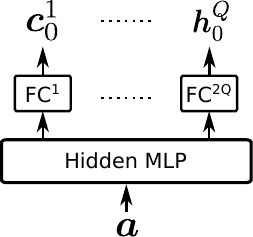}
	\caption{Init Model. FC = Fully Connected Layer.}
	\label{fig-init}
\end{wrapfigure}
The Init Model $f_{init}$, produces the initial state $\boldsymbol{C}_0$ of the LSTM-Stack. $f_{init}$ is intended to `look' at the entire image ($\boldsymbol{a}$) and setup the decoder appropriately before it starts generating the output. 
\begin{IEEEeqnarray}{rCl}
	f_{init} & : & \boldsymbol{a}  \rightarrow  ( \boldsymbol{c}^1_{0}, \ldots , \boldsymbol{c}^Q_{0}, \, \boldsymbol{h}^1_{0}, \ldots , \boldsymbol{h}^Q_{0} ) \IEEEeqnarraynumspace \\
	&& \boldsymbol{h}_0^q, \boldsymbol{c}_0^q \in \mathbb{R}^n \nonumber
\end{IEEEeqnarray}
That said, since it only provides a very small improvement in performance in exchange for over 7 million parameters, its need could be questioned. $f_{init}$ is modeled as an MLP with common hidden layers and $2Q$ distinct output layers, one for each element of $C_0$, connected as in Figure \ref{fig-init}. See the \suppmat ~for more detail and discussion.

\subsection{Training}
\label{training}
\begin{wrapfigure}[5]{r}{0.45\textwidth}
	\vspace{-18pt}
\begin{IEEEeqnarray}{rCl}
	\mathcal{J} &=& -\frac{1}{\tau} {log} \left( P_r \left( \boldsymbol{y}|\boldsymbol{a} \right)  \right) + \lambda_R \mathcal{R} \IEEEeqnarraynumspace \IEEEyesnumber \label{eqn-J} \\
	\mathcal{R} &=& \frac{1}{2} \sum_{\theta} \theta^2   \IEEEyessubnumber
\end{IEEEeqnarray}
\end{wrapfigure}
The entire model was trained end-to-end by minimizing the objective function $\mathcal{J}$ (Equation \ref{eqn-J}) using back propagation through time.
The first term in Equation \ref{eqn-J} is the average (per-word) log perplexity of the predicted sequence\footnote{i.e. Average cross-entropy, negative log-likelihood or negative log-probability.} and is the main objective. 
$\mathcal{R}$ is the L2-regularization term, equal to L2-norm of the model's parameters $\theta$ (weights and biases) and $\lambda_R$ is a hyperparameter requiring tuning. Following \citet{Xu2015ShowAA} at first, we had included a penalty term intended to bias the distribution of the cumulative attention placed on an image-location ${\alpha}_l := \sum_{t=1}^{\tau}\alpha_{t,l}$. However we removed it for various reasons which are discussed in the \suppmat ~along with other details and analyses.

We split the dataset into two fixed parts: 1) training dataset = 90-95\% of the data and 2) test dataset 5-10\%. At the beginning of each run, 5\% of the training dataset was randomly held out as the validation-set and the remainder was used for training. Therefore, each such run had a different training/validation data-split, thus naturally cross-validating our learnings across the duration of the project. We trained the model in minibatches of 56 using the ADAM optimizer \cite{Kingma2014AdamAM}; periodically evaluating it over the validation set\footnote{Evaluation cycle was run once or twice per epoch and/or when a training BLEU score calculated on sequences decoded using CTC-Decoding\cite{Graves2006ConnectionistTC} jumped significantly.}. For efficiency we batched the data such that each minibatch had similar length samples. For the final evaluation however, we fixed the training and validation dataset split and retrained our models for about 100 epochs ($\sim 2\frac{1}{2}$ days). We then picked the model-snapshots with the best validation BLEU score and evaluated the model over the test-dataset for publication. Table \ref{table-training1} lists the training parameters and metrics of various configurations. Training sequence predictions ($\hat{\boldsymbol{y}}$) were obtained by CTC-decoding  \cite{Graves2006ConnectionistTC} $\boldsymbol{p}_t$. Training BLEU score was then calculated over 100 consecutive mini-batches. We used two Nvidia GeForce 1080Ti graphics cards in a \href{https://github.com/petewarden/tensorflow_makefile/blob/master/tensorflow/models/image/cifar10/cifar10_multi_gpu_train.py}{parallel towers configuration}. Our \href{\implementation}{implementation} uses the \href{https://www.tensorflow.org/}{Tensorflow toolkit} and is distributed under AGPL license. 
\begin{table*}[!hbtp]
	\caption{Training metrics. $\lambda_R=0.00005 \text{~and~} \beta_2 = 0.9$ for all runs. The number after @ sign is the training epoch of the selected model-snapshot. $^*$ denotes that the row corresponds to Table \ref{table-scores}.}
	\begin{tabular}{lll|lll|llll}
		\hline
		\textbf{Dataset} & \textbf{Model} & \textbf{Init}  &\textbf{$\beta_1$}  & \textbf{Train}  & \textbf{Train} & \textbf{Validation}  & \textbf{Valid'n}\\
		                 &                & \textbf{Model?}&                    & \textbf{Epochs} & \textbf{BLEU}  & \textbf{BLEU}        & \textbf{ED}        \\
		\hline 
		I2L-140K    & I2L-STRIPS & Yes & 0.5 & 104 & 0.9361 & 0.8900@$72^*$ & 0.0677 \\
				    & I2L-STRIPS & No  & 0.5 & 75  & 0.9300 & 0.8874@$62$   & 0.0691 \\
					& I2L-NOPOOL & Yes & 0.5 & 104 & 0.9333 & 0.8909@$72^*$ & 0.0684 \\
					& I2L-NOPOOL & No  & 0.1 & 119 & 0.9348 & 0.8820@$92$   & 0.0738 \\
		\hline
		Im2latex-90k& I2L-STRIPS & Yes & 0.5 & 110 & 0.9366 & 0.8886@$77^*$ & 0.0688 \\
		            & I2L-STRIPS & No  & 0.5 & 161 & 0.9386 & 0.8810@$118$  & 0.0750 \\
		\hline
	\end{tabular}
	\centering
	\label{table-training1}
\end{table*}

\section{Results}
\begin{figure*}
	\begin{tabular}{llcc}
		& Input Image / Rendered Sequence & $\boldsymbol{y}_{len}$ & $\boldsymbol{\hat{y}}_{len}$ \\
		\hline
		\input{strips_matched_strs2}
	\end{tabular}
	\centering
	\caption[A Sample of Correct Predictions]{A sample of correct predictions by \textsc{i2l-strips}. We've shown the long predictions hence lengths are touching 150. Note that at times the target length is greater than the predicted length and at times the reverse is true (though the original and predicted images were identical). All such cases would evaluate to a less than perfect BLEU score or edit-distance. This happens in about 40\% of the cases. For more examples visit \href{\website}{our website}.}
	\label{fig-good-preds}
\end{figure*}
\begin{figure*}
	\begin{tabular}{lll}
		& $\boldsymbol{y}$ & $\boldsymbol{\hat{y}}$ \\
		\hline
		\input{strips_unmatched_preds2}
	\end{tabular}
	\centering
	\caption[Random Sample of Mistakes]{A random sample of mistakes made by \textsc{i2l-strips}. Observe that usually the model gets most of the formula right and the mistake is only in a small portion of the overall formula (e.g. sample \# 1; generating one subscript $_t$ instead of an $_l$). In some cases the mistake is in the font and in some cases the images are identical but were incorrectly flagged by the image-match evaluation software (e.g. sample \# 0 \& \#17). In some cases the predicted formula appears more correct than the original! (sample \# 10 where position of the subscript $_{\beta}$ has been `corrected' by \textsc{i2l-strips}).
	}
	\label{fig-bad-preds}
\end{figure*}
Given that there are multiple possible \LaTeX sequences that will render the same math image, ideally we should perform a visual evaluation. However, since there is no widely accepted visual evaluation metric, we report corpus BLEU (1,2,3 \& 4 grams) and per-word Levenstein Edit Distance\footnote{i.e. Edit distance divided by number of words in the target sequence.} scores (see Table \ref{table-scores}). We also report a (non-standard) exact visual match score\footnotemark[103] which reports the percentage of exact visual matches, discarding all partial matches. While the predicted and targeted images match in at least 70\%\footnotemark[103] of the cases, the model generates different but correct sequences (i.e. $\boldsymbol{y} \neq \boldsymbol{\hat{y}}$) in about 40\% of the cases (Figure~\ref{fig-good-preds}). For the cases where the images do not exactly match, the differences in most cases are minor (Figure~\ref{fig-bad-preds}). Overall, our models produce syntactically correct sequences\footnote{i.e. Those that were successfully rendered by \LaTeXe.} for at least 99.85\% of the test samples (Table \ref{table-scores}). Please visit our website to see hundreds of sample visualizations, analyses and discussions, data-set and source-code.
\begin{table*}[!hbtp]
	\caption{Test results. Im2latex-100k results are from \citet{Deng2017ImagetoMarkupGW}. The last column is the percentage of successfully rendering predictions.}
	\begin{tabular}{llllll}
		\hline
		\textbf{Dataset} & \textbf{Model} & \textbf{BLEU}  & \textbf{Edit}     & \textbf{Visual} & \textbf{Compiling} \\
		&         & \textbf{Score} & \textbf{Distance} & \textbf{Match\footnotemark[103]} & \textbf{Predictions} \\
		\hline
		I2L-140K & \textsc{i2l-nopool} & \textbf{89.0}\% &  0.0676 & 70.37\% & 99.94\% \\
		& \textsc{i2l-strips} & \textbf{89.0}\% & 0.0671 & 69.24\%  & 99.85\% \\
		\hline
		Im2latex-90k & \textsc{i2l-strips} & \textbf{88.19}\%& 0.0725 & 68.03\%  & 99.81\% \\
		Im2latex-100k      & \textsc{Im2Tex}     & 87.73\%         & - & \textbf{79.88}\%  & - \\
		\hline
	\end{tabular}
	\centering
	\label{table-scores}
\end{table*}
\footnotetext[103]{We use the 'match without whitespace' algorithm provided by \citet{Deng2017ImagetoMarkupGW} wherein two images count as matched if they match pixel-wise discarding white columns and allowing for upto 5 pixel image translation (a pdflatex quirk). It outputs a binary match/no-match verdict for each sample - i.e. partial matches however close, are considered a non-match.}

\subsection{Model Interpretability via attention} Since the LSTM stack only sees a filtered view (i.e. focal-region) of the input, it can only base its predictions on the focal-regions seen thus far and initial-state $C_0$. Further since the init-model has a negligible impact on performance we can drop it from the model (Table \ref{table-training1}) and thereby the dependency on $C_0$ (now randomly initialized). Therefore if $\boldsymbol{I}_t$ is the focal-region at step $t$ defined by the predicate $\boldsymbol{\alpha}_{t,l} > 0$, then $ \boldsymbol{p}_t \left(\hat{\boldsymbol{y}}_t \right) = f_L\left( \boldsymbol{I}_t, \boldsymbol{I}_{t-1} \ldots \boldsymbol{I}_0 \right)$ where $f_L$ represents the LSTM-stack and Deep Output Layer. This fact aids considerably in interpreting the predictions of the model. We found heat-map type visuals of the focal-regions (Figure \ref{fig-att}) very useful in interpreting the model even as we were developing it.
\paragraph{Object detection via attention:}
Additionally, we observe that the model settles on a step-by-step alignment of $\boldsymbol{I}_t$ with the output-word's location on the image: i.e. $\boldsymbol{p}_t \left(\hat{\boldsymbol{y}}_t \right) \approx f_L\left( \boldsymbol{I}_t \right)$. In other words $\boldsymbol{I}_t$ marks the bounding-box of $\hat{\boldsymbol{y}}_t$ even though we trained without any bounding-box data. Therefore our model -whose encoder has a narrow receptive field- can be applied to the object detection task without requiring bounding box training data, bottom-up region proposals or pretrained classifiers. Note that this is not possible with encoder architectures having wide receptive fields, e.g. those that employ a RNN \cite{Deng2017ImagetoMarkupGW, Bahdanau2014NeuralMT} because their receptive fields encompass the entire input. A future work will quantify the accuracy of object detection \cite{Liu2017AttentionCI} using more granular receptive fields. \citet{DBLP:journals/corr/PedersoliLSV16} have also used attention for object detection but their model is more complex in that it specifically models bounding-boxes although it doesn't require them for training.

\subsection{Dataset}
\label{dataset}
Datasets were created from single-line \LaTeX ~math formulas extracted from scientific papers and subsequently processed as follows:
\begin{enumerate*}[label=\arabic*)]
	\item Normalize the formulas to minimize spurious ambiguity.\footnote{Normalization was performed using the method and software used by \cite{Deng2017ImagetoMarkupGW} which parses the formulas into an AST and then converts them back to normalized sequences.}
	\item Render the normalized formulas using pdflatex and discard ones that didn't compile or render successfully.
	\item Remove duplicates.
	\item Remove formulas with low-frequencey words (frequency-threshold = 24 for Im2latex-90k and 50 for I2l-140K).
	\item Remove images bigger than $1086 \times 126$ and formulas longer than 150.
\end{enumerate*}
\footnotetext[104]{\href{https://zenodo.org/record/56198\#.WnzcT3UbMQ9}{Im2latex-100k dataset} is provided by \cite{Deng2017ImagetoMarkupGW}.}
Processing the \href{https://zenodo.org/record/56198\#.WnzcT3UbMQ9}{Im2latex-100k dataset}\footnotemark[104] (103559 samples) as above resulted in the \textbf{Im2latex-90k} dataset which has 93741 samples. Of these, 4648 were set aside as the test dataset and the remaining 89093 were split into training (95\%) and validation (5\%) sets before each run (section \ref{training}).
We found the {Im2latex-90k} dataset too small for good generalization and therefore augmented it with additional samples from \href{http://www.cs.cornell.edu/projects/kddcup/datasets.html}{KDD Cup 2003}. This resulted in the \textbf{I2L-140K} dataset with 114406 (training), 14280 (validation) and 14280 (test) samples. Since the normalized formulas are already space separated token sequences, no additional tokenization step was necessary. The vocabulary was therefore produced by simply identifying the set of unique space-separated words in the dataset.
\paragraph{Ancillary material}
All ancillary material: Both datasets, our model and data-processing source code, visualizations, result samples etc. is available at \href{\website}{our website}. \Suppmat ~is provided alongside this paper.

\nocite{Xu2015ShowAA}
\nocite{Deng2017ImagetoMarkupGW}
\nocite{Bluche2014ACO}

\bibliographystyle{apalike}
\bibliography{I2LPaper_NIPS2018}

\appendix
\begin{figure*}[p]
	\begin{tabular}{lll}
		& $\boldsymbol{y}$ & $\boldsymbol{\hat{y}}$ \\
		\hline
		\input{strips_rand_sample}
	\end{tabular}
	\centering
	\caption[Random Sample of Predictions]{A random sample of predictions of \textsc{i2l-strips} containing both good and bad predictions. Note that though this is a random sample, prediction mistakes are not obvious and it takes some effort to point them out! For more examples visit \href{\website}{our website.}\footnotemark[105]}
	\label{fig-rand-preds}
\end{figure*}

\section{Qualitative analyses and details}
\label{observations}
This section is an appendix to the paper. We present here further details, analyses and discussion of our experiments and comparison with related work.

\subsection{Encoder}
\label{encoder-commentary}
Table \ref{table-cnn} shows the configuration of the Encoder CNN. All convolution kernels have shape (3,3), stride (1,1) and $tanh$ non-linearity, whereas all maxpooling windows have shape (2,2) and stride (2,2).
\begin{table}[!hbtp]
	\caption{Specification of the Encoder CNN.}
	\begin{tabular}{lcc}
		\hline
		{Layer} & Output Shape & {Channels} \\
		\hline
		Input (Image) & $128 \times 1088$ & 1\\
		Convolution & $128 \times 1088$ & 64 \\
		Maxpool & $64 \times 544$ & 64 \\
		Convolution) & $64 \times 544$ & 128 \\
		Maxpool & $32 \times 272$ & 128 \\
		Convolution & $32 \times 272$ & 256 \\
		Maxpool & $16 \times 136$ & 256 \\
		Convolution & $16 \times 136$ & 512 \\
		Maxpool & $8 \times 68$ & 512 \\
		Convolution & $8 \times 68$ & 512 \\
		Maxpool & $4 \times 34 = (\acute{H} \times \acute{W})$ & $512 = (\acute{D})$ 
	\end{tabular}
	\centering
	\label{table-cnn}
\end{table}
We initially experimented with the output of the VGG16 model \cite{Simonyan2014VeryDC} - per \citet{Xu2015ShowAA}. However (presumably since VGG16 was trained on a different dataset and a different problem) the BLEU score didn't improve beyond 40\%. Then we started training VGG16 along with our model but the end-to-end model didn't even start learning (the log-loss curve was flat) - possibly due to the large overall depth of the end-to-end model. Reducing the number of convolution layers to 6 and changing the non-linearity to $tanh$ (to keep the activations in check) got us good results. Further reducing number of layers to 5 yielded the same performance, therefore we stuck with that configuration (Table \ref{table-cnn}). In additon, we experimented with \textsc{i2l-strips} because it reduces the rectangular image-map to a linear map, thereby presumably making the alignment model's task easier because now it would only need to scan in one-dimension. However, it performed around the same as \textsc{i2l-nopool} and therefore that hypothesis was debunked. In fact we prefer I2L-NOPOOL since it has fewer parameters and its attention model has sharper focal-regions which helps with model interpretation.

\subsection{Attention model}
\label{att-analyses}
\begin{wraptable}{r}{0.45\textwidth}
	\caption[Visual Attention MLP]{Specification of the Visual Attention Model MLP. L = 34 for I2L-STRIPS and and 136 for I2L-NOPOOL.}
	\begin{tabular}{lll}
		\textbf{Layer} & \textbf{Num Units} & \textbf{Activation}\\
		\hline
		3 (output) & L & softmax \\
		2 & max(128, L) & tanh \\
		1 & max(256, L) & tanh
	\end{tabular}
	\centering
	\label{table-att}
\end{wraptable}
Table \ref{table-att} specifies the configuration of the attention model MLP. \citealt{Xu2015ShowAA}'s formulation of attention model ($\alpha_{t,l} = MLP \left( \boldsymbol{a}_l ; \, \boldsymbol{H}_{t-1}  \right)$) receives inputs from only a single image location. In comparison, our formulation ($\boldsymbol{\alpha}_t =  f_{att} \left( \boldsymbol{a} ; \, \boldsymbol{H}_{t-1} \right)$) receives the full encoded image $\boldsymbol{a}$ in its input. This change was needed because the previous formulation did not progress beyond a point, presumably because this problem warranted a wider receptive field. The new formulation works equally well with different pooling strides (and correspondingly different values of L).

Also, \citealt{Xu2015ShowAA}'s formulation of $\boldsymbol{z}_t =  \beta_t \cdot \boldsymbol{\alpha}_t \cdot \boldsymbol{a}$ includes a scalar $\beta_t = MLP(\boldsymbol{H}_{t-1})$ which informs the LSTM how much emphasis to place on the image v/s the language model. Experimentally we found that it had no impact on end-to-end performance, therefore we dropped it from our model.

\citealt{Xu2015ShowAA} also use a simpler formula for $\mathcal{A} = \sum_{l=1}^{L} \left( \sum_{t=1}^{\tau}\alpha_{t,l} -1 \right)^2$ which they call `doubly stochastic optimization'. Our formulation uses the true mean of $\alpha_l$, $\tau / L$ instead of 1, normalizes it to a fixed range so that it can be compared across models and more importantly, includes a target-ASE term $ASE_T$. Without this term, i.e. with $ASE_T=0$, $\mathcal{A}$ would bias the attention model towards uniformly scanning all the L image locations. This is undesirable since there are many empty regions of the images where it makes no sense for the attention model to spend much time. Conversely, there are some densely populated regions (e.g. a symbol with complex superscript and subscripts) where the model would reasonably spend more time because it would have to produce a longer output sequence. In other words, the optimal scanning pattern would have to be non-uniform - $ASE_T \neq 0$. Also, the scanning pattern would vary from sample to sample, but $ASE_T$ is set to a single value (even if zero) for all samples. Therefore we preferred to remove the attention-model bias altogether from the objective function by setting $\lambda_A=0$ in all situations except when the attention model needed a 'nudge' in order to `get off the ground'. In such cases we set $ASE_T$ based on observed values of $ASE_N$ (Table \ref{table-training2}).


\subsection{LSTM stack}
\label{lstm-comments}
\begin{figure}[!h]
	\begin{minipage}{0.4\textwidth}
		\centering
		\includegraphics[width=\linewidth]{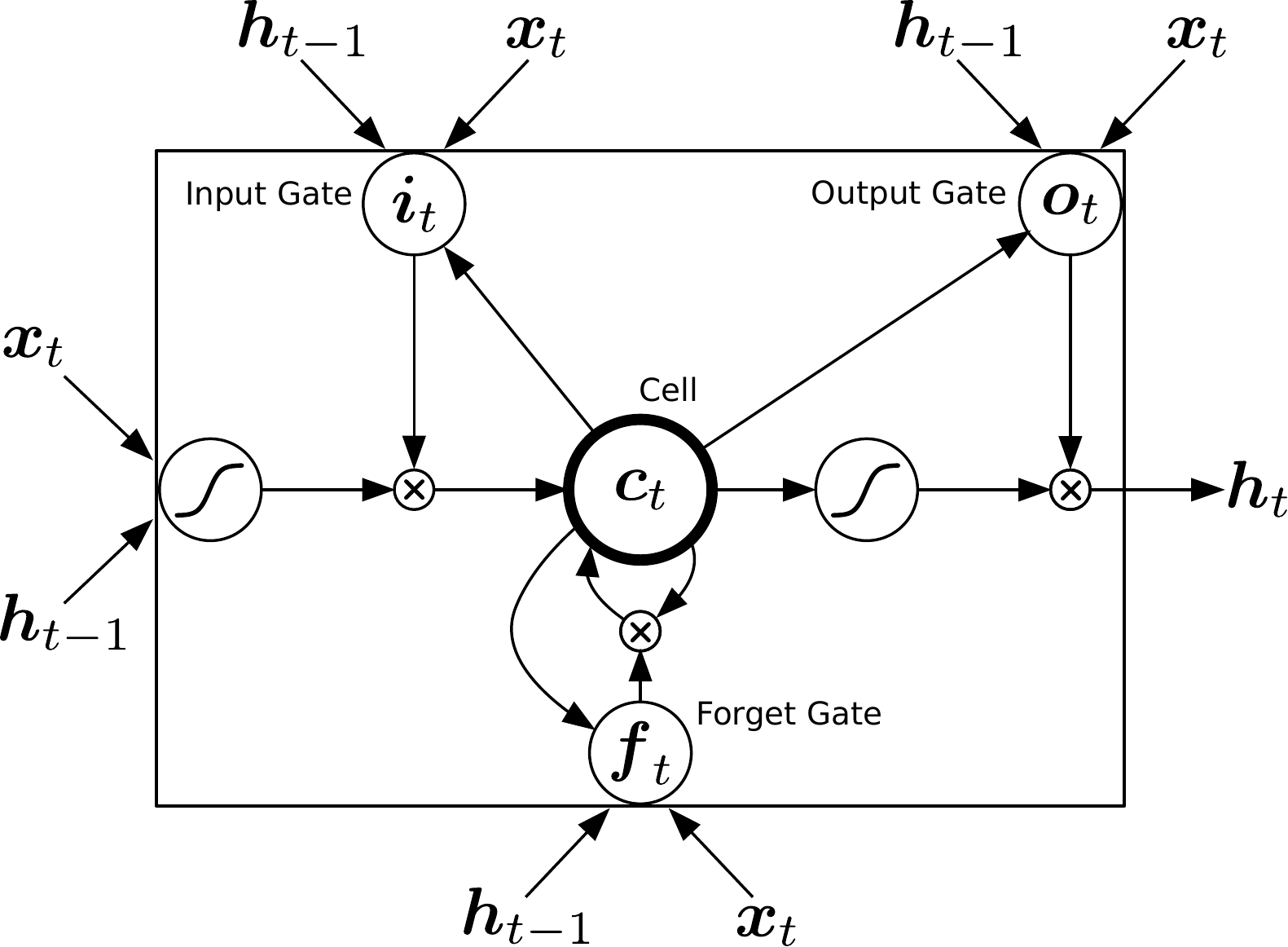}
		\caption[LSTM]{LSTM Cell}
		\label{fig-lstm}
	\end{minipage}
	\begin{minipage}{0.6\textwidth}
		\begin{IEEEeqnarray}{rCl}
			\itm &=& \sigma\left(\wtmat{x}{i} \xtm + \wtmat{h}{i} \htm[t-1] + \wtmat{c}{i} \ctm[t-1]  + \boldsymbol{b}_i\right) \nonumber \\
			\ftm &=& \sigma\left(\wtmat{x}{f} \xtm + \wtmat{h}{f} \htm[t-1] + \wtmat{c}{f} \ctm[t-1] + \boldsymbol{b}_f \right) \nonumber \\
			\ctm &=& \ftm \ctm[t-1] + \itm \tanh \left(\wtmat{x}{c} \xtm + \wtmat{h}{c} \htm[t-1] + \boldsymbol{b}_c \right) \nonumber \\
			\otm &=& \sigma\left(\wtmat{x}{o} \xtm + \wtmat{h}{o} \htm[t-1] + \wtmat{c}{o} \ctm + \boldsymbol{b}_o\right) \nonumber \\
			\htm &=& \otm \tanh(\ctm) \nonumber \\
			&& \itm, \ftm, \otm, \ctm, \htm \in \mathbb{R}^n \label{eqn-lstm}
		\end{IEEEeqnarray}
	\end{minipage}
\end{figure}
Our LSTM cell implementation (Figure. \ref{fig-lstm} and equation \ref{eqn-lstm}) follows \citet{DBLP:journals/corr/abs-1303-5778, Zaremba2014RecurrentNN}.
In equation \ref{eqn-lstm} $\sigma$ is the logistic sigmoid function and $\itm$, $\ftm$, $\otm$, $\ctm$ and $\htm$ are respectively the \emph{input gate}, \emph{forget gate}, \emph{output gate}, \emph{cell} and \emph{hidden} activation vectors of size $n$.

During experimentation our penultimate LSTM-stack which had 3 LSTM layers with 1000 units each, gave us a validation score of 87.45\%. At that point experimental observations suggested that the LSTM stack was the accuracy 'bottleneck' because other sub-models were performing very well. Increasing the number of LSTM units to 1500 got us better validation score - but a worse overfit. Reducing the number of layers down to 2 got us the best overall validation score. In comparison, \citet{Xu2015ShowAA} have used a single LSTM layer with 1000 cells.

\subsection{Deep output layer}
\label{output-comments}
\begin{wraptable}{r}{0.5\textwidth}
	\caption[Output Layer Configuration]{Configuration of the Deep Output Layer MLP. $K$ = 339 and 358 for I2L-140K and Im2latex-90k datasets respectively.}
	\begin{tabular}{lll}
		\textbf{Layer} & \textbf{Num Units} & \textbf{Activation}\\
		\hline
		3 (output) & K & softmax \\
		2 & max(358, K) & tanh \\
		1 & max(358, K) & tanh
	\end{tabular}
	\centering
	\label{table-output-layer}
\end{wraptable}
Note that the output layer receives skip connections from the LSTM-Stack input ($\boldsymbol{p}_t = f_{out}(\boldsymbol{H}_t; \, \boldsymbol{z}_t; \, \boldsymbol{Ey}_{t-1})$). We observed a ~2\% impact on the BLEU score with the addition of input-to-output skip-connections. This leads us to believe that adding skip-connections within the LSTM-stack may help further improve model accuracy. Overall accuracy also improved by increasing the number of layers from 2 to 3. Lastly, observe that this sub-model is different from \citet{Xu2015ShowAA} wherein the three inputs are affine-transformed into $D$ dimensions, summed and then passed through one fully-connected layer. After experimenting with their model we ultimately chose to instead feed the inputs (concatenated) to a fully-connected layer thereby allowing the MLP to naturally learn the input-to-output function. We also increased the number of layers to 3, changed activation function of hidden units from relu to tanh\footnotemark[101] and ensured that each layer had at least as many units as the softmax layer ($K$).
\footnotetext[101]{We changed from relu to tanh partly in order to remedy `activation-explosions' which were causing floating-point overflow errors.}
\subsection{Init model}
\label{init-model-comments}
\begin{wraptable}{}{0.5\textwidth}
	\caption{Init Model layers.}
	\begin{tabular}{llll}
		\textbf{Layer} & \textbf{Num} & \textbf{Units} & \textbf{Activation} \\
		&&&\textbf{Function}\\
		\hline
		Output & 2Q & n & tanh \\
		Hidden & 1 & 100 & tanh
	\end{tabular}
	\centering
	\label{table-init-model}
\end{wraptable}
The init model MLP is specified in Table \ref{table-init-model}. We questioned the need for the Init Model and experimented just using zero values for the initial state. That caused a slight but consistent decline ($<$ 1\%) in the validation score, indicating that the initial state learnt by our Initial State Model did contribute in some way towards learning and generalization. Note however that our Init Model is different than \citealp{Xu2015ShowAA}, in that our version uses all $L$ feature vectors of $\boldsymbol{a}$ while theirs takes the average. We also added a hidden layer and used $tanh$ activation function instead of $relu$. We did start off with their version but that did not provide an appreciable impact to the bottom line (validation). This made us hypothesize that perhaps taking an average of the feature vectors was causing a loss of information; and we mitigated that by taking in all the $L$ feature vectors without summing them. After making all these changes, the Init Model yields a consistent albiet small performance improvement (Table. \ref{table-init-efficacy}). But given that it consumes $\sim$7.5 million parameters, its usefulness remains in question.
\begin{table}[h]
	\caption{Impact of the Init Model on overall performance. Since it comprises 10-12\% of the total params, it may as well be omitted in exchange for a small performance hit.}
	\begin{tabular}{llll}
		\textbf{Model} & \textbf{Init Model} & \textbf{Validation} & \textbf{Num}\\
		& \textbf{Present?} & \textbf{BLEU} & \textbf{Params} \\
		\hline
		\textsc{i2l-nopool} & Yes & 89.09\% & 7,569,300 \\
		\textsc{i2l-nopool} & No & 88.20\% & 0\\
		\textsc{i2l-strips} & Yes & 89.00\% & 7,569,300 \\
		\textsc{i2l-strips} & No & 88.74\% & 0
	\end{tabular}
	\centering
	\label{table-init-efficacy}
\end{table}

\subsection{Training and dataset}
\subsubsection{Alpha penalty}
Please see equations \ref{eqn-J2} through \ref{eqn-alpha-l2}. The loss function equation stated in the paper is Equation \ref{eqn-J2} but with $\lambda_A$ set to 0. That was the case when training models who's results we have published, however at other times we had included a penalty term $\lambda_A \mathcal{A}$ which we discuss next. Observe that while {$\sum_{l}^{L} \alpha_{t,l} = 1 $}, there is no constraint on how the attention is distributed across the $L$ locations of the image. The term $\lambda_{A}\mathcal{A}$ serves to steer the variance of $\alpha_l$ by penalizing any deviation from a desired value. ${ASE}$ (Alpha Squared Error) is the sum of squared-difference between $\alpha_l$ and its mean $\tau/L$; and $ASE_N$ is its normalized value \footnote{It can be shown that  $\tau^2 \left( \frac{L-1}{L} \right)$ is the maximum possible value of $ASE$.} $\in$ [0,100]\footnote{We normalize $ASE$ so that it may be compared across batches, runs and models.}. Therefore $ASE_N \propto ASE \propto \sigma_{\alpha_l}^2$.  $ASE_T$ which is the desired value of $ASE_N$, is a hyperparameter that needs to be discovered through experimentation\footnote{Start with $ASE_T=0$, observe where $ASE_N$ settles after training, then set $ASE_T$ to that value and repeat until approximate convergence.}. Table \ref{table-training2} shows training results with alpha-penalty details.
\begin{table*}[!hbtp]
	\caption{Training metrics. $\lambda_R=0.00005 \text{~and~} \beta_2 = 0.9$ for all runs.}
	\begin{tabular}{lll|lll|llll}
		\hline
		\textbf{Dataset} & \textbf{Model} & \textbf{Init}  & \textbf{$\lambda_A$}  &\textbf{$\beta_1$}  & \textbf{Training}  & \textbf{Training}   & \textbf{Validation} & ${\overline{ASE_N}}$\\
		                 &                & \textbf{Model?}&                       &                    & \textbf{Epochs}    & \textbf{BLEU}       & \textbf{ED}         & \\
		\hline 
		I2L-140K    & I2L-STRIPS & Yes & 0.0    & 0.5 & 104 & 0.9361 & 0.0677 & 5.3827 \\
				    & I2L-STRIPS & No  & 0.0    & 0.5 & 75  & 0.9300 & 0.0691 & 4.9899\\
					& I2L-NOPOOL & Yes & 0.0    & 0.5 & 104 & 0.9333 & 0.0684 & 4.5801\\
					& I2L-NOPOOL & No  & 0.0    & 0.1 & 119 & 0.9348 & 0.0738 & 4.7099\\ 
		\hline
		Im2latex-90k& I2L-STRIPS & Yes & 0.0    & 0.5 & 110 & 0.9366 & 0.0688 & 5.1237\\
		& I2L-STRIPS & No  & 0.0005 & 0.5 & 161 & 0.9386 & 0.0750 & 4.8291\\
		\hline
	\end{tabular}
	\centering
	\label{table-training2}
\end{table*}
\begin{wrapfigure}{r}{0.5\textwidth}
	\vspace{-15pt}
	\begin{IEEEeqnarray}{rCl}
		\mathcal{J} &=& -\frac{1}{\tau} {log} \left( P_r \left( \boldsymbol{y}|\boldsymbol{a} \right)  \right) + \lambda_R \mathcal{R} + \lambda_{A} \mathcal{A} \IEEEeqnarraynumspace \IEEEyesnumber \label{eqn-J2} \\
		\mathcal{R} &=& \frac{1}{2} \sum_{\theta} \theta^2   \IEEEyessubnumber  \\
		\mathcal{A} &=& \left(  {ASE}_{N} - {ASE}_T \right)  \IEEEyessubnumber   \\
		{ASE}_N &=& \frac{100}{ \tau^2 \left( \frac{L-1}{L} \right) } \cdot ASE  \IEEEyessubnumber \label{eqn-ASE_N2} \\
		{ASE} &=& { \sum_{l=1}^{L} \left( \alpha_l - \frac{\tau}{L} \right)^2 }  \IEEEyessubnumber  \\
		\alpha_l &:=& \sum_{t=1}^{\tau}\alpha_{t,l} \IEEEyessubnumber \label{eqn-alpha-l2}
	\end{IEEEeqnarray}
\end{wrapfigure}
Default values of $\beta_1 \text{and} \beta_2$ of the ADAM optimizer - 0.9 and 0.99 - yielded very choppy validation score curves with frequent down-spikes where the validation score would fall to very low levels, ultimately resulting in lower peak scores. Reducing the first and second moments (i.e. $\beta_1 \text{and} \beta_2$) fixed the problem suggesting that the default momentum was too high for our `terrain'. We did not use dropout for regularization, however increasing the data-set size (I2L-140K) and raising the minimum-word-frequency threshold from 24 (Im2latex-90k) to 50 ((I2L-140K)) did yield better generalization and overall test scores (Table \ref{table-training2}). Finally, normalizing the data\footnote{Normalization was performed using the method and software used by \cite{Deng2017ImagetoMarkupGW} which parses the formulas into an AST and then converts them back to normalized sequences.} yielded about 25\% more accuracy than without.

\end{document}